\documentclass[11pt,a4paper]{article}
\usepackage[hyperref]{acl2019}
\usepackage{times}
\pdfoutput=1
\usepackage{latexsym}
\usepackage{comment}
\usepackage{todonotes}
\usepackage{graphicx}
\usepackage{diagbox}
\usepackage{multirow}
\usepackage{makecell}
 \usepackage{textcomp}
 \usepackage{lscape}
 \usepackage{longtable}
\def\hlinewd#1{%
  \noalign{\ifnum0=`}\fi\hrule \@height #1 \futurelet
   \reserved@a\@xhline}
\usepackage{url}

\aclfinalcopy 

\title{REflex: Flexible Framework for Relation Extraction in Multiple Domains}

\author{Geeticka Chauhan \\
  MIT CSAIL \\
  \texttt{geeticka@mit.edu} \\\And
  Matthew B. A. McDermott \\
  MIT CSAIL \\
  \texttt{mmd@mit.edu} \\\And
  Peter Szolovits \\
  MIT CSAIL \\
  \texttt{psz@mit.edu}\\}

\date{May 9, 2019}

\begin{document}
\maketitle
\begin{abstract}

 Systematic comparison of methods for relation extraction (RE) is difficult because many experiments in the field are not described precisely enough to be completely reproducible and many papers fail to report ablation studies that would highlight the relative contributions of their various combined techniques. In this work, we build a unifying framework for RE, applying this on three highly used datasets (from the general, biomedical and clinical domains) with the ability to be extendable to new datasets. By performing a systematic exploration of modeling, pre-processing and training methodologies, we find that choices of pre-processing are a large contributor performance and that omission of such information can further hinder fair comparison. Other insights from our exploration allow us to provide recommendations for future research in this area. 
 
 
 
\end{abstract}


\section{Introduction}
Relation Extraction (RE) has gained a lot of interest from the community with the introduction of the Semeval tasks from 2007 by \citep{girju2007semeval} and 2010 by \citep{hendrickx2009semeval}. The task is a subset of information extraction (IE) with the goal of finding semantic relationships between concepts in a given sentence, and is an important component of Natural Language Understanding (NLU). Applications include automatic knowledge base creation, question answering, as well as analysis of unstructured text data. Since the introduction of RE tasks in the general and medical domains, many researchers have explored the performance of different neural network architectures on the datasets \cite{socher2012semantic, zeng2014relation, liu2016drug, sahu2016clinical}.

However, progress in RE is hampered by reproducibility issues as well as the difficulty in assessing which techniques in the literature will generalize to novel tasks, datasets and contexts. To assess the extent of these problems, we performed a manual review of 53 relevant neural RE papers\footnote{The 53 papers were filtered from a list of 728 papers skimmed for relevance. Appendix \ref{appendix:quant-lit-review} contains paper details. } citing the three datasets \cite{hendrickx2009semeval,segura2013semeval,uzuner20112010}. The procedure for finding these papers is highlighted in \cite{geeticka2019REflex}. 
 
\paragraph{Reproducibility} Reproducibility is important for validating previous work and building upon it \cite{fokkens2013offspring}. Lack of reproducibility can be attributed to many factors such as difficulty in availability of source code \cite{ince2012case} and omission of sources of variability such as hyperparameter details \cite{claesen2015hyperparameter}. We found that only 16 out of the 53 relevant papers
 had released their source code. 14 out of 53 papers were evaluated on multiple datasets, but the source code was publicly available for only five of those. Despite this, much of this code was lacking in modularity to be easily extendable to new datasets. In many cases, the process of reproducing the paper results was often unclear and lack of documentation made this more difficult. Even though most papers mentioned some hyperparameter details, important details were missing such as number of epochs, batch size, random initialization seed, if any, and details about early stop if that technique was applied. 

 \paragraph{Ablation Studies} Lack of generalizability is caused by a dearth of appropriate empirical evaluation to identify the source of modeling gains. Ablation studies are important for identifying sources of improvements in results. Among the 53 papers that we looked at, 20 of the 24 papers in the general domain performed ablation studies. However, only 10 out of 29 papers in the medical domain performed one. Among these ablation studies, key details related to pre-processing were missing, which we found critical in our experiments.

In the absence of such information about causes of large variability of results, \textit{fair comparison} of models becomes difficult. In this paper, we present an open-source unifying framework enabling the comparison of various training methodologies, pre-processing, modeling techniques, and evaluation metrics. The code is available at \url{https://github.com/geetickachauhan/relation-extraction}. 

The experimental goals of this framework are identification of sources of variability in results for the three datasets and provide the field with a strong baseline model to compare against for future improvements. The design goals of this framework are identification of best practices for relation extraction and to be a guide for approaching new datasets. 

By performing systematic comparison on three datasets, we find that 1) pre-processing choices can cause the largest variations in performance, 2) reporting scores on one test set split is problematic due to split bias. We perform other analyses in section \ref{sec:discussion} and also include recommendations for future research in this field in section \ref{sec:conclusion}. 

Upon testing various combinations of our approaches, we achieve results near state of the art ranges for the three datasets: 85.89\% macro F1 for Semeval 2010 task 8 dataset \cite{hendrickx2009semeval} i.e. \texttt{semeval}, 71.97\% macro F1 for DDI Extraction 2013 \cite{segura2013semeval} i.e. \texttt{ddi} and 71.01\% micro F1 for i2b2/VA 2010 relation classification dataset \cite{uzuner20112010} i.e. \texttt{i2b2}. We refer to \texttt{ddi} and \texttt{i2b2} as medical datasets, as they belong to the biomedical and clinical domains, respectively.

\begin{table}[b]
    \centering
     \scalebox{0.9}{
    \begin{tabular}{l l l l l}
    \hline \hline 
         \textbf{Dataset} & \textbf{Rel} & \textbf{Eval}  & \textbf{Agreement} & \textbf{Det}\\ \hline
         \texttt{semeval} & 18 & Macro &  0.6-0.95 & No\\
         \texttt{ddi} & 5 & Macro & \textgreater0.8; 0.55-0.72 & Yes \\
         \texttt{i2b2} & 8 & Micro & -  & Yes \\ \hline \hline 
    \end{tabular}
    }
    \caption{Dataset information, with columns Rel = number of relations, Eval = evaluation metric (all F1 scores), Agreement = Inter-annotator agreement, Det = whether detection task from section \ref{subsec:methodology-evaluation} was evaluated on. Rel column only includes relations used in official evaluation metric. \texttt{ddi} was built from two separately annotated sources and therefore contains two inter-annotator agreements. }
    \label{tab:datasets}
\end{table}

\section{Datasets}
\label{sec:datasets}

We summarize important information about these datasets in table \ref{tab:datasets}. We introduce \textit{detection} and \textit{classification} tasks in section \ref{subsec:methodology-evaluation}, but also indicate the tasks evaluated for each dataset in table \ref{tab:datasets}.


\paragraph{Semeval 2010} \texttt{semeval} consists of 8000 training sentences and 2,717 test sentences for the multi-way classification of semantic relations between pairs of nominals. Not included in the official evaluation is an \textit{Other} class which is considered noisy, with annotators choosing this class if no fit was found in the other classes. It is important to note that this is a synthetically generated dataset, and \textit{detection} scores were not calculated due to the noisy nature of the \textit{Other} class.   


\paragraph{DDI Extraction} \texttt{ddi} consists of 1,017 texts with 18,491 pharmacological substances and 5,021 drug-drug interactions from Pubmed articles in the pharmacological literature. \textit{None} class indicating no interaction between the drug pairs is included in the evaluation metric calculation. 

\paragraph{i2b2/VA 2010 relations} \texttt{i2b2} consists of discharge summaries from Partners Healthcare and the MIMIC II Database \cite{saeed2011multiparameter}. They released 394 training reports, 477 test reports and 877 unannotated reports. After the challenge, only a part of the data was publicly released for research. \textit{None} relation was present in the data and not considered in the official evaluation. 



\section{Methodology}
\label{sec:methodology}

Our framework breaks up processing into different stages, allowing for future modular addition of components. First, a \texttt{formatter} converts the raw dataset into a common comma separated value (CSV) input format accepted by the \texttt{pre-processor}, and this information is then fed to the \texttt{model}, which performs the training, after which \texttt{evaluation} is performed on the test set. With our framework, we test the following variations in the main components:

\subsection{Pre-Processing}
We test various pre-processing methods after performing simple tokenization and lower-casing of the words: entity blinding used by \citet{liu2016drug}, stop-word and punctuation removal, and digit normalization commonly applied for \texttt{ddi} in \cite{zhao2016drug}, and named entity recognition related replacement (we call this NER blinding). We used the spaCy framework\footnote{https://github.com/explosion/spaCy} for tokenization and to identify punctuation and digits. 

Entity blinding and NER blinding are similar concept blinding techniques where the first is performed based on gold standard annotations, while the second is performed by running NER on the original sentence. We replace the words in the sentence matching the entity or named entity span with the target label and use those for training and testing.

Entity labels for \texttt{semeval} were not annotated with type information, whereas \texttt{ddi} identified drugs and \texttt{i2b2} identified medical problems, tests and treatments. Therefore, entity labels for \texttt{semeval} were \textit{ENTITY}, for \texttt{ddi} were \textit{DRUG} and for \texttt{i2b2} were \textit{PROBLEM}, \textit{TREATMENT} and \textit{TEST}. In this paper, we use \textit{fine-grained concept type} to refer to the presence of more than one concept type, as in the the case of \texttt{i2b2}. 

NER labels for \texttt{semeval} consisted of those provided by the large english model by spaCy and provided standard types such as \textit{PERSON} and \textit{ORGANIZATION}, whereas those for the medical datasets was provided by the ScispaCy medium size model and did not provide types \cite{neumann2019scispacy}. In this case, blinding consisted of replacing the words in the sentence by \textit{Entity}. 

We chose the spaCy model for NER to complement the extendable design goals of \texttt{REflex}. Other options such as cTAKES \cite{savova2010mayo} for clinical data and MetaMAP\footnote{https://metamap.nlm.nih.gov} for biomedical data are highly specific to the dataset type and require running additional scripts outside of the \texttt{REflex} pipeline.  

\subsection{Modeling}
\label{subsec:methodology-modeling}

We employ a baseline model based upon \cite{zeng2014relation}, \cite{santos2015classifying} and \cite{jin2018semeval}, which is a convolutional neural network (CNN) with position embeddings and a ranking loss (referred to as \texttt{CRCNN} in this paper). We initialize the model with pre-trained word embeddings: the senna embeddings by \citet{collobert2011natural} for the general domain dataset and the \texttt{PubMed-PMC-wikipedia} embeddings released by \citet{moen2013distributional} for the medical domain. We test several perturbations on top of \texttt{CRCNN} model, such as piecewise max-pooling, as suggested by \citet{zeng2015distant} and the more recent ELMo embeddings by \citet{peters2018elmo}. To compare different featurizations of contextualized embeddings, we also employ the embeddings generated by the BERT model (rather than the standard fine-tuning approach). For ELMo, we use the Original (5.5B) model weights in \texttt{semeval} and PubMed contributed model weights in the medical datasets released by \cite{peters2018elmo}. For BERT, we use the BERT-large uncased model (without whole word masking) in \texttt{semeval} released by \cite{devlin2018bert}, BioBERT
by \cite{lee2019biobert} in \texttt{ddi} and Clinical BERT by \cite{alsentzer2019publicly} in \texttt{i2b2}. 

The fine-tuning approach, which tends to be computationally expensive, has been thoroughly explored for multiple tasks, including medical relation extraction by \citet{lee2019biobert}, but the approach of featurizing them with an existing model has not been explored in the literature as much. We tested different ways of featurizing the BERT contextualized embeddings for researchers who want to utilize a less computationally intensive technique, while still aiming for performance gains for their task.

Because ELMo provides token level embeddings, we chose to concatenate them with the word and position embeddings from \texttt{CRCNN} before the convolution phase. However, BERT provides word-piece level as well as sentence level embeddings. The first was concatenated similar to ELMo (which we call BERT-tokens), while the second was concatenated with the fixed size sentence representation outputted after convolution of word and position embeddings (BERT-CLS). 

\subsection{Training}
We explore two ways of doing hyperparameter tuning: manual tuning and random search \cite{bergstra2012random}. 

Evaluating on three datasets meant that we needed to identify a default list of hyperparameters by tuning on one of the datasets before we could identify the hyperparameter list for the other two. We chose \texttt{semeval} for initial tuning due to its larger literature and because the \texttt{CRCNN} model was originally evaluated on this dataset. We started with reference hyperparameters listed in \citet{zeng2014relation} and \citet{santos2015classifying} and identified default hyperparameters after tuning on a dev set randomly sampled from the training data of the \texttt{semeval} dataset. These default hyperparameters\footnote{listed in source code} were used as starting points for manual tuning on the medical datasets as well as random search for all datasets. 

We perform manual tuning on a subset of the hyperparameters, mentioned in table \ref{tab:manual-hyperparam-search}. In order to avoid overfitting in cross validation pointed out by \citet{cawley2010over}, we perform a nested cross validation procedure, keeping a dev fold for hyperparameter tuning and a held out fold for score reporting. 

On these dev folds, we perform paired t-tests for each of the perturbations to the parameters listed in table \ref{tab:manual-hyperparam-search}. Our first pass involves changing one hyperparameter per experiment and noting the ones that cause a statistically significant improvement, which helps us identify a narrower list of hyperparameters to tune on. We further refine the hyperparameter values in our second pass by testing on values similar to those that were leading to statistically significant improvements in the first pass. For example, if we noticed that lower epoch values were helpful in the first pass, we tested them in combination with the other optimal hyperparameter values (from first pass) in the second pass. 

For each of the datasets, we tuned based on their official challenge evaluation metrics listed in section \ref{sec:datasets}. \texttt{ddi} and \texttt{i2b2} had 5-fold nested cross validation performed on them, whereas \texttt{semeval} had 10-fold cross validation performed. 


Random search was performed based on the official evaluation metrics for each dataset, on a fixed dev set randomly sampled from the training data. Final distributions are listed in table \ref{tab:random-hyperparam-search}.


\subsection{Evaluation}
\label{subsec:methodology-evaluation}

The official challenge problems for all datasets compared models based on multi-class classification, but for the medical datasets, we were also interested in looking at the changes in model performance if we treated the task as a binary classification problem. This was based on the rationale that in the drug literature, for example, pharmacologists would not want to sacrifice the ability to identify a potentially life threatening drug interaction pair, even if the type of the drug pair is not known. Therefore, we report results for both multi-class and binary classification scenarios. For clarity, we refer to them in the rest of the paper as \textit{classification} and \textit{detection} respectively.

\textit{Detection} results were obtained using our evaluation scripts by treating existing relations as one class, ignoring the types outputted by the model. The other class in this task was the \textit{None} or \textit{Other} class, representing non-existing relations. Note that we did not re-train our model for this.

In addition to evaluating on two tasks for the medical and one task for the general dataset, we comment on the implications of different evaluation metrics in section \ref{subsec:discussion-evaluation}. 
\section{Results}
\label{sec:results}

For experiments on the medical datasets i.e. \texttt{i2b2} and \texttt{ddi}, we used hyperparameters found from manual search individually performed on them. \texttt{semeval} had the default hyperparameters used for its experiments. These sets of hyperparameters were used in all experiments other than those reported in table \ref{tab:hyperparameter}, where we compare hyperparameter tuning methodologies.  

Once we had a fixed set of hyperparameters for each dataset, we tested the perturbations for pre-processing as well as modeling in tables \ref{tab:preprocessing} and \ref{tab:modeling}. Perturbations on the hyperparameter search are listed in table \ref{tab:hyperparameter} and compare performance with different hyperparameter values found using different tuning strategies. 

We generate the standard \textit{classification} and the additional \textit{detection} scores by the procedure described in section \ref{subsec:methodology-evaluation}, and report these results under the \textit{Class} and \textit{Detect} columns.


We also report additional experiments in tables \ref{tab:i2b2-additional} and \ref{tab:ddi-additional} based on the improvements found in tables \ref{tab:preprocessing} and \ref{tab:modeling}. For all results tables, we report official test set results at the top, with accompanying cross validated results (averaged over all folds with their standard deviation) in smaller font below them.\footnote{Results tables for metrics other than the official ones were omitted in the interest of space, but their analysis exists in section \ref{subsec:discussion-evaluation}.}

\begin{table}[t]
    \centering
     \scalebox{0.9}{
    \begin{tabular}{l r}
    \hline \hline
         \textbf{Hyperparameter} & \textbf{Values}  \\ \hline
         epoch & \{50,100,150,200\} \\ 
         lr decay &  [1e-3, 1e-4, 1e-5] \\
         sgd momentum & \{T, F\} \\
         early stop & \{T, F\} \\
         pos embed & \{10, 50, 80, 100\} \\
         filter dimension & \{50, 150\} \\
         filter size & {2-3-4, 3-4-5} \\
         batch size & \{70, 30\} \\ \hline \hline
    \end{tabular}
    }
    \caption{Hyperparameters explored for the first pass of manual search. lr decay means learning rate decay at [60, 120] epochs, pos embed refers to the position embedding size.}
    \label{tab:manual-hyperparam-search}
\end{table}

\begin{table}[t]
    \centering
     \scalebox{0.9}{
    \begin{tabular}{l r}
    \hline \hline
         \textbf{Hyperparameter} & \textbf{Distributions}  \\ \hline
         epoch & uniform(70, 300) \\
         lr & \{constant, decay\} \\
         lr init & \small uniform(1e-5, 0.001) \\ \hline
         \multirow{2}{*}{filter size} & 2-3, 2-3-4, 2-3-4-5\\
                                      & 3-4-5, 3-4-5-6 \\ \hline
         early stop & \{T, F\} \\
         batch size & uniform(30, 70) \\ \hline \hline
    \end{tabular}
    }
    \caption{Hyperparameter distributions for random search. Those written in \{\} are picked with equal probabilities. The learning rate (lr) was uniformly initialized, and decayed from 0.001 to the intialized value at half of the number of epochs. If early stop was true, patience was set to a fifth of the number of epochs. We ran 100-120 experiments for each dataset to search for optimal hyperparameters.}
    \label{tab:random-hyperparam-search}
\end{table}


 \begin{table*}[t]
     \centering
      \scalebox{0.9}{
     \begin{tabular}{l|c|c c | c c }
     \hline \hline
          \multirow{2}{*}{\diagbox[width=10em]{Preprocess}{Dataset}} & \texttt{semeval} & \multicolumn{2}{c|}{\texttt{ddi} }& \multicolumn{2}{c}{\texttt{i2b2} }             \\ 
                                                &                               & Class                         & Detect                       & Class                         & Detect           \\ \hline
          \multirow{2}{*}{Original}             & \textbf{81.55}                & 65.53                         & 81.74                         & 59.75                         & 83.17             \\ 
                                                & \small 80.85 (1.31)           & \small 82.23 (0.32)           & \small 88.40 (0.48)           & \small 70.10 (0.85)           & \small 86.45 (0.58)     \\ \hline
          \multirow{2}{*}{Entity Blinding}      & 72.73                         & \textbf{67.02}                & \textbf{82.37}                & \textbf{68.76}                & \textbf{84.37} \\ 
                                                & \small 71.31 (1.14)           & \small 83.56 (2.05)\textbullet& \small 89.45 (1.05)\textbullet& \small 76.59 (1.07)           & \small 88.41 (0.37)  \\ \hline
          \multirow{2}{*}{Punct and Digit}      & 81.23                         & 63.41                         & 80.49                         & 58.85                         & 81.96         \\ 
                                                & \small 80.95 (1.21)\textbullet& \small 80.44 (1.77)           & \small 87.52 (0.98)           & \small 69.37 (1.43)\textbullet& \small 85.82 (0.43)     \\ \hline
          \multirow{2}{*}{Punct, Digit and Stop}& 72.92                         & 55.87                         & 76.57                         & 56.19                         & 80.47         \\ 
                                                & \small 71.61 (1.25)           & \small 78.52 (1.99)           & \small 85.65 (1.21)           & \small 68.14 (2.05)\textbullet& \small 84.84 (0.77) \\ \hline
          \multirow{2}{*}{NER Blinding}         & 81.63                         & 57.22                         & 79.03                         & 50.41                         & 81.61 \\
                                                & \small 80.85 (1.07)\textbullet& \small 78.06 (1.45)           & \small 86.79 (0.65)           & \small 66.26 (2.44)           & \small 86.72 (0.57)\textbullet \\ \hline \hline
          
     \end{tabular}
     }
     \caption{Pre-processing techniques with \texttt{CRCNN} model. Row labels Original = simple tokenization and lower casing of words, Punct = punctuation removal, Digit = digit removal and Stop = stop word removal. Test set results at the top with cross validated results (average with standard deviation) below. All cross validated results are statistically significant compared to Original pre-processing ($p < 0.05$) using a paired t-test except those marked with a \textbullet}
     \label{tab:preprocessing}
 \end{table*}

 \begin{table*}[t]
     \centering
      \scalebox{0.9}{
     \begin{tabular}{l|c|c c | c c }
     \hline \hline
          \multirow{2}{*}{\diagbox[width=10em]{Modeling}{Dataset}}      & \texttt{semeval} & \multicolumn{2}{c|}{\texttt{ddi} } & \multicolumn{2}{c}{\texttt{i2b2} }            \\ 
                                                &                               & Class                         & Detect                        & Class                 & Detect           \\ \hline
          \multirow{2}{*}{CRCNN}                & 81.55                         & 65.53                         & 81.74                         & 59.75                 & 83.17             \\ 
                                                & \small 80.85 (1.31)           & \small 82.23 (0.32)           & \small 88.40 (0.48)           & \small 70.10 (0.85)   & \small 86.45 (0.58)     \\ \hline
          \multirow{2}{*}{Piecewise pool}       & 81.59                         & 63.01                         &  80.62                        & 60.85                 &  83.69             \\ 
                                                & \small 80.55 (0.99)\textbullet& \small 81.99 (0.38)\textbullet& \small 88.47 (0.48)\textbullet& \small 73.79 (0.97)   & \small 89.29 (0.61)  \\ \hline
          \multirow{2}{*}{BERT-tokens}          &  85.67                        &  \textbf{71.97}               &  \textbf{86.53}               &  63.11                & \textbf{84.91}                \\ 
                                                & \small 85.63 (0.83)           & \small 85.35 (0.53)           & \small 90.70 (0.46)           & \small 72.06 (1.36)   & \small 87.57 (0.75)           \\ \hline
          \multirow{2}{*}{BERT-CLS}             &  82.42                        & 61.3                          & 79.63                         & 56.79                 & 81.91          \\ 
                                                & \small 80.83 (1.18)\textbullet& \small 82.71 (0.68)\textbullet& \small 88.35 (0.77)\textbullet& \small 67.37 (1.08)   & \small 85.43 (0.36)           \\ \hline
          \multirow{2}{*}{ELMo}                 &   \textbf{85.89}              & 66.63                         & 83.05                         & \textbf{63.18}        & 84.54                   \\
                                                & \small 84.79 (1.08)           & \small 84.53 (0.96)           & \small 90.11 (0.56)           & \small 72.53 (0.80)   & \small 87.81 (0.34)    \\ \hline \hline
          
     \end{tabular}
     }
     \caption{Modeling techniques with original pre-processing. Test set results at the top with cross validated results (average with standard deviation) below. All cross validated results are statistically significant compared to \texttt{CRCNN} model ($p < 0.05$) using a paired t-test except those marked with a \textbullet. In terms of statistical significance, comparing contextualized embeddings with each other reveals that BERT-tokens is equivalent to ELMo for \texttt{i2b2}, but for \texttt{semeval} BERT-tokens is better than ELMo and for \texttt{ddi} BERT-tokens is better than ELMo only for detection.}
     \label{tab:modeling}
 \end{table*}


  \begin{table*}[t]
     \centering
      \scalebox{0.9}{
     \begin{tabular}{l|c|c c | c c }
     \hline \hline
          \multirow{2}{*}{\diagbox[width=13em]{Hyperparam Tuning}{Dataset}}      & \texttt{semeval} & \multicolumn{2}{c|}{\texttt{ddi} } & \multicolumn{2}{c}{\texttt{i2b2} }             \\ 
                                                &                               & Class                         & Detect                        & Class                 & Detect           \\ \hline
          \multirow{2}{*}{Default}              &  81.55                        &  62.55                        &  80.29                        &  55.15                & 81.98                     \\ 
                                                & \small 80.85 (1.31)           & \small 81.62 (1.35)           & \small 87.76 (1.03)           & \small 67.28 (1.83)   & \small 86.57 (0.58)           \\ \hline
          \multirow{2}{*}{Manual Search}        & -                             & \textbf{65.53}                & \textbf{81.74}                & \textbf{59.75}        & \textbf{83.17}             \\ 
                                                & \small                        & \small 82.23 (0.32)\textbullet& \small 88.40 (0.48)\textbullet& \small 70.10 (0.85)   & \small 86.45 (0.58)\textbullet   \\ \hline
          \multirow{2}{*}{Random Search}        & \textbf{82.2}                 & 62.29                         &  79.04                        & 55.0                  &  80.77             \\ 
                                                & \small 81.10 (1.26)\textbullet& \small 75.43 (1.48)           & \small 83.54 (0.60)           & \small 60.66 (1.43)   & \small 82.73 (0.49)  \\ \hline \hline

     \end{tabular}
     }
     \caption{Hyperparameter tuning methods with original pre-processing and fixed \texttt{CRCNN} model. Test set results at the top with cross validated results (average with standard deviation) below. All cross validated results are statistically significant compared to Default with $p<0.05$ except those marked with a \textbullet. Note that hyperparameter tuning can involve much higher performance variation depending on the distribution of the data. Therefore, even though there is no statistical significance in the manual search case for the held out fold in the ddi dataset, there was statistical significance for the dev fold which drove those set of hyperparameters. For both ddi and i2b2 datasets, manual search is better than random search with $p < 0.05$. }
     \label{tab:hyperparameter}
 \end{table*}
 
   \begin{table}[h]
     \centering
     \scalebox{0.9}{
     \begin{tabular}{l c c}
     \hline \hline
          \diagbox[width=7em]{Technique}{Task}      & Classification & Detection     \\ \hline 
          \multirow{2}{*}{E + ent }            &  70.46                &  86.17                     \\ 
                                                & \small 77.70(1.26)   & \small 89.36 (0.50)   \\ \hline
          \multirow{2}{*}{B + ent }            &  70.56                &  85.66                     \\ 
                                                & \small 76.72 (1.04)   & \small 88.63 (0.33)   \\ \hline
          \multirow{2}{*}{E + piece + ent }        &  70.62                    & 86.14         \\ 
                                                & \small  79.41 (0.53)              & \small 90.37 (0.44)  \\ \hline
          \multirow{2}{*}{B + piece + ent }     &  \textbf{71.01}    & \textbf{86.26}         \\ 
                                                & \small 79.51 (1.09)               & \small 90.34 (0.53)  \\ \hline
          \multirow{2}{*}{piece + ent }        & 69.73         & 85.44               \\ 
                                                & \small 78.12 (1.10)  & \small 89.74 (0.44)  \\ \hline
        \multirow{2}{*}{E + piece}        & 63.19         & 84.92               \\ 
                                                & \small 74.76 (0.68)  & \small 89.90 (0.37)  \\ \hline
        \multirow{2}{*}{B + piece}        & 63.23         & 85.45               \\ 
                                                & \small 74.67 (0.89)  & \small 89.61 (0.68)  \\ \hline \hline

     \end{tabular}
     }
     \caption{Additional experiments for \texttt{i2b2}. E = ELMo, B = BERT-tokens, ent  = entity blinding, piece = piecewise pooling. All results are statistically significant compared to BERT-tokens and ELMo models respectively from table \ref{tab:modeling} and piece + ent row is statistically significant compared to piecewise pool model as well as entity blinding model. These are all statistically significantly better than the \texttt{CRCNN} model from table \ref{tab:modeling} }
     \label{tab:i2b2-additional}
 \end{table}
 
  \begin{table}[h]
     \centering
      \scalebox{0.9}{
     \begin{tabular}{l c c}
     \hline \hline
          \diagbox[width=7em]{Technique}{Task}      & Classification & Detection     \\ \hline 
          \multirow{2}{*}{E + ent }            &  68.69                &  83.72     \\ 
                                                & \small 86.25 (1.54)   & \small 91.35 (0.90)   \\ \hline
          \multirow{2}{*}{B + ent }            &  \textbf{70.66}                &  \textbf{85.35}                     \\ 
                                                & \small 85.79 (1.54)   & \small 91.26 (0.63)   \\ \hline \hline

     \end{tabular}
     }
     \caption{Additional experiments for \texttt{ddi}. E = ELMo, B = BERT-tokens, ent  = entity blinding. Results are not statistically significant compared to BERT-tokens and ELMo models respectively from table \ref{tab:modeling} and not from each other either. }
     \label{tab:ddi-additional}
 \end{table}

\section{Discussion}
\label{sec:discussion}

Recently, CNNs have achieved strong performance for text classification and are typically more efficient than recurrent architectures \cite{bai2018empirical, kalchbrenner2014CNN, wang2015semantic, zhang2015character}. The speed of our baseline \texttt{CRCNN} model allows us to explore multiple alternatives for every stage of our pipeline. We discuss these results pertaining to the \textit{classification} task for all datasets and the \textit{detection} task for the medical datasets.


\subsection{Pre-processing} 

Often, papers fail to mention the importance of pre-processing in performance improvements. Experiments in table \ref{tab:preprocessing} reveal that they can cause larger variations in performance than modeling. 

We applied pre-processing changes with the \texttt{CRCNN} model with default hyperparameters for \texttt{semeval} and manual hyperparameters for the medical datasets. All comparisons are performed against the original pre-processing technique, which involved using the original dataset sentences in training and test.


Punctuation and digits hold more importance for the \texttt{ddi} dataset, which is a biomedical dataset, compared to the other two datasets. We looked at examples where this technique led to an incorrect prediction, but original pre-processing led to a correct one to investigate the source of performance further. The examples indicate that removal of punctuation is driving worse performance compared to the normalization of digits. A detailed analysis for these is present in \cite{geeticka2019REflex}.

Stop word removal is a common technique in Natural Language Processing (NLP) to simplify the sentence by cutting out commonly used words such as \textit{the} and \textit{is} in order to simplify the sentence. We found that stop words seem to be important for relation extraction for all three datasets that we looked at, to a smaller degree for \texttt{i2b2} compared to the other two datasets. Looking at examples misclassified by this technique revealed important stop words for different relations, which indicates that the removal of stop words is not beneficial in the relation extraction setting. Example types are shown in \cite{geeticka2019REflex}. 

The availability of fine-grained concept types is likely to boost performance in relation extraction settings. The \texttt{i2b2} dataset provided fine-grained concept types in the form of medical problem, test and treatments. Entity blinding causes almost 9\% improvement in \textit{classification} performance and 1\% improvement in \textit{detection} performance. In contrast, \texttt{ddi} only provided gold standard annotations for drug types in the sentence, and while this does not cause statistically significant improvements for cross validation, it does improve test set classification performance by about 1.5\% and detection performance by 1\%. For these medical datasets, NER blinding consisted of replacing the detected named entities by \textit{Entity} because named entity types were not available. Due to the coarse-grained nature of the entities, it hurts \textit{classification} performance significantly, and \textit{detection} performance a little. 

While entity blinding hurts performance for \texttt{semeval}, possibly due to the coarse-grained nature of the replacement, NER blinding does not hurt performance. Looking at misclassified examples for entity blinding and NER blinding techniques supports this hypothesis \cite{geeticka2019REflex}. 

To recall, entity blinding involved replacement of entity words by \textit{Entity}, while NER blinding involved replacing named entities in the sentence with labels such as \textit{ORGANIZATION} and \textit{PERSON}. In settings where fine-grained entity blinding may not be helping, they may be helpful as added features into the model, as shown by \cite{socher2012semantic}. 

For the medical datasets, while \textit{classification} performance varies highly with different pre-processing techniques, \textit{detection} is relatively unaffected. In a setting where one cares more about detection of relationships rather than multi-class classification, one would be able to get away with using non-complicated pre-processing techniques to maintain reasonable performance. 

\subsection{Split Bias} All three datasets evaluate models based on one score on the test set, which is common practice for NLP challenges. Reporting one score as opposed to a distribution of scores has been shown to be problematic by \citet{reimers2017reporting} for sequence tagging. Recently, \citet{crane2018questionable} discuss similar problems for question-answering. We show that even if you keep the same random initialization seed (all our experiments have a fixed random initialization seed), train-test set split bias can be another source of variation in scores. 

In our experiments, significance testing of some cross validated results reveals no significance even when the test set result improves in performance. This is particularly concerning for \texttt{ddi} where entity blinding (called drug blinding in the literature) is used as a standard pre-processing technique without ablation studies demonstrating its effectiveness. Our results suggest the contrary: entity blinding seems to help test set performance for \texttt{ddi} in table \ref{tab:preprocessing}, but shows no statistical significance. Table \ref{tab:ddi-additional} further shows that using this in conjunction with other techniques results in test score variations despite being statistically insignificant. 

No statistical significance is seen even when the test set result worsens in performance for BERT-CLS and Piecewise Pool in table \ref{tab:modeling} where it hurts test set performance on \texttt{ddi} but is not statistically significant when cross validation is performed. BERT-CLS improves test set result for \texttt{semeval} but is not found to be statistically significant.


\subsection{Modeling} 

In table \ref{tab:modeling}, we tested the generalizability of the commonly used piecewise pooling technique proposed in \cite{zeng2015distant}, a variant of which was applied in the model by \citeauthor{luo2017segment} for \texttt{i2b2}. We also tested the improvements offered by different featurizations of contextualized embeddings, which has not been explored much for relation extraction. 

Modeling changes were applied with the original pre-processing technique for the \texttt{CRCNN} model with default hyperparameters for \texttt{semeval} and manual hyperparameters for the medical datasets. All comparisons are performed with the baseline performance of the \texttt{CRCNN} model. 

While piecewise pooling helps \texttt{i2b2} by 1\%, it hurts test set performance on \texttt{ddi} and doesn't affect performance on \texttt{semeval}. While it may be intuitive to split pooling by entity location, this technique is not generalizable to other datasets.

We also found that while contextualized embeddings generally boost performance, they should be concatenated with the word embeddings before the convolution stage to cause a significant boost in performance. We found ELMo and BERT-tokens to boost performance significantly for all datasets, but that BERT-CLS hurt performance for the medical datasets. While BERT-CLS boosted test set performance for \texttt{semeval}, this was not found to be a statistically significant difference for cross validation. Note that we featurized ELMo similarly to BERT-tokens and the details are present in section \ref{subsec:methodology-modeling}. 

This indicates that the technique of featurizing the contextualized embeddings is important for a CNN architecture. Concatenating the contextualized embeddings with the word embeddings keeps a tighter coupling, which is helpful for relation extraction where the word-level ordering might be essential in predicting the relation type.


\subsection{Hyperparameter Tuning} 
\citet{bergstra2012random} show the superiority of random search over grid search in terms of faster convergence, but leave to future work automating the procedure of manual tuning, i.e. sequential optimization. Bayesian optimization strategies could help with this \cite{snoek2012practical} but often require expert knowledge for correct application. We tested how manual tuning, requiring less expert knowledge than Bayesian optimization, would compare to the random search strategy in table \ref{tab:hyperparameter}. For both \texttt{i2b2} and \texttt{ddi} corpora, manual search outperformed random search. 



\subsection{Evaluation Metrics}
\label{subsec:discussion-evaluation}

Picking the right evaluation metric for a dataset is critical, and it is important to choose a metric that has the biggest delta between different model performances for example types we care about. Tables for different metric results for all datasets are provided in Appendix \ref{appendix:evaluation-metric}. 

When using micro and macro statistics (precision, recall and F1), class imbalance dictates the one to pick. Macro statistics are highly affected by imbalance, whereas micro statistics are able to recover well. Despite suffering due to class imbalance, though, macro statistics may be more appropriate than micro as they provide stronger discriminative capabilities by providing equal importance to classes of smaller sizes.  However, micro statistics are as discriminative as macro statistics in settings when the classes are relatively balanced. We are going to talk about the \textit{classification} tasks in the next two paragraphs.

Compared to \texttt{semeval}, \texttt{ddi} and \texttt{i2b2} suffer from stark class imbalances. \texttt{semeval} has a number of examples in non-\textit{Other} classes ranging from 200 or 300 to 1000. \textit{Other} class has about 3000 examples which are not included in the official metric calculations. \texttt{ddi} has one class with 228 examples, while the others have about 1000 examples. The \textit{None} class has 21,948 examples which is included for the official score calculations. \texttt{i2b2} has five classes in the 100-500 range, while the others contain about 2000 examples. \textit{None} is the largest class with 19,934 examples. 

Using micro statistics is reasonable for \texttt{i2b2} because the highly imbalanced class is not included in the calculations. Therefore, this metric is able to be as discriminative as macro statistics. For example, test set micro F1 between baseline and entity blinding techniques is 59.75 and 68.76, while that for macro F1 is 36.44 and 43.76. In contrast, using micro statistics is a bad idea for \texttt{ddi} because the performance on the \textit{None} class would drive most of the predictive results of the model. For example, micro-F1 between baseline and NER blinding is 88.69 and 86.18, whereas macro-F1 is 65.53 and 57.22. \texttt{semeval} does not have a stark contrast between micro and macro scores due to \textit{Other} class not being included in the calculation. Using either metric to evaluate models is reasonable for this dataset. 

The detection task does not suffer from such variations due to the lower class imbalance. For example, \texttt{ddi} dataset micro-F1 between baseline and NER blinding model is 90.01 and 88.74, while macro-F1 is 81.74 and 79.03. This further suggests that modeling differences and pre-processing differences cause more variation in performance in settings when the class imbalance is higher. 




\section{Comparison with SOTA}
\label{sec:comparison-sota}

The best \textit{classification} test set results found are listed in table \ref{tab:best-results}. Note that we do not compare the \textit{extraction} task for datasets other than \texttt{ddi} because the official challenges only compared \textit{classification} results. Even though the official challenge did not rank models based on the \textit{detection} task, recent papers in the \texttt{ddi} literature mention these results.

 \begin{table}[h]
     \centering
     \begin{tabular}{l c c}
     \hline \hline
          Dataset           & Result    & Technique     \\ \hline 
          \texttt{semeval}  &  85.89    & E     \\ 
          \texttt{ddi}      & 71.97, 86.53 & B \\
          \texttt{i2b2}     & 71.01     & B + piece + ent \\ \hline \hline
     \end{tabular}
     \caption{Best test set \textit{classification} results for all datasets, except \texttt{ddi} where \textit{detection} results are mentioned after the classification results. piece = Piecewise pooling, ent = entity blinding, E = ELMo, B = BERT-tokens. Result corresponds to F1 scores, macro for \texttt{semeval} and \texttt{ddi}, but micro for \texttt{i2b2}. }
     \label{tab:best-results}
 \end{table}

\citet{wang2016relation} report a result of 88\% on \texttt{semeval} and do not provide any public source code for replication purposes. Despite being below the state of the art range, \texttt{REflex} provides the best performing publicly available model for this dataset. \citet{zheng2017attention} report the best result on \texttt{ddi} (77.3\%) but perform negative instance filtering, which is a highly specific pre-processing technique that does not fit with the flexible nature of \texttt{REflex}. This technique cuts specific examples from the dataset, but the paper is unclear about whether train as well as test data are shortened. If the test data is being shortened, the performance comparison becomes unfair due to evaluation on different test samples. Unfortunately, source code was not publicly available to answer these questions.

Note that \citet{zhao2016drug} show that negative instance filtering causes a 4.1\% improvement in test set performance. If \texttt{REflex} were to use this pre-processing technique, it would reach close to the state-of-the-art (SOTA) number on the \textit{classification} task. On the other hand, results from the \textit{detection} results \textit{outperform} this model by 2.53\%. 

\citet{sahu2016clinical} (code unavailable) report a state of the art result of 71.16\% on \texttt{i2b2}, which the results in table \ref{tab:best-results} are able to match. Note that \cite{rink2011automatic} report a result of 73.7\% with a support vector machine, but they used a larger version of the dataset. Comparison against different subsets of the dataset would not be fair. 

Comparison against these numbers demonstrates that \texttt{REflex} is the only open-source framework, providing performance near SOTA ranges for the three datasets. Therefore, \texttt{REflex} can be used as a strong baseline model in future relation extraction studies.

\section{Conclusion}
\label{sec:conclusion}

Our findings reveal variations offered by pre-processing and training methodologies, which often go unreported. They indicate that comparing models without having these techniques standardized can make it difficult to assess the true source of performance gains. Our key findings are:
 
 1. Pre-processing can have a strong effect on performance, sometimes more than modeling techniques, as is the case of \texttt{i2b2}. Concept types seem to offer useful information, perhaps revealing more general semantic information in the sentence that can help with predictions. Fine-grained Gold standard annotated concept types are most beneficial, but those from automatically extracted packages may also be useful as long as they consist of multiple types. Punctuation and digits may hold more importance in biomedical settings, but stop words hold significance in all settings.


2. Reporting on one test set score can be problematic due to split bias, and a cross validation approach with significance tests may help ease some of this bias. Drug blinding for \texttt{ddi} is commonly used in the literature but does not seem to offer any statistically significant improvements. Therefore, it is unnecessary to use in this domain.

3. Contextualized embeddings are generally helpful but the featurizing technique is important: for CNN models, concatenating them with the word embeddings before convolution is most beneficial.

4. Picking the right hyperparameters for a dataset is important to performance. We suggest an initial manual hyperparameter search based on cross validation significance tests because that may be sufficient in most cases. If one is not pressed for time, random search is a reasonable automated option for hyperparameter tuning, but requires more experience for picking the right search space and the right distributions for the hyperparameters.

5. Picking the right evaluation metrics for a new dataset should be driven by class imbalance issues for the classes chosen to be evaluated on. 

  
\section*{Acknowledgments}
This work was funded in part by a collaborative agreement between MIT and Wistron Corp, the National Institutes of Health (National Institutes of Mental Health grant P50-MH106933), and a Mitacs Globalink Research Award.
Finally, the authors would like to thank Di Jin and Elena Sergeeva from the MIT-CSAIL Clinical Decision Making Group for providing helpful feedback. 

\bibliographystyle{acl_natbib}
\bibliography{final}

\newpage
\clearpage
\onecolumn
\appendix

\begin{landscape}
\section{Quantitative Literature Review}
\label{appendix:quant-lit-review}

  \begin{longtable}{l l l l l l l l}
     \hline \hline
          \textbf{paper}      & \textbf{cite} & \textbf{code} & \textbf{ablation} 
          & \textbf{hyperparam} & \textbf{cross val} & \textbf{word-embed} & \textbf{datasets}    \\ \hline 
        \endfirsthead
        \hline \hline
        \textbf{Paper}      & \textbf{cite} & \textbf{code} & \textbf{ablation} 
          & \textbf{hyperparam} & \textbf{cross val} & \textbf{word-embed} & \textbf{datasets}    \\ \hline 
        \endhead
 \hline 
        \citet{socher2012semantic} & 890 & y & \textbullet & y & \textbullet & y & 2  \\
        
 \hline 
        \citet{zeng2014relation} & 477 & \textbullet & y & y & y & y & 1  \\
        
 \hline 
        \citet{santos2015classifying} & 220 & \textbullet & y & y & y & y & 1  \\
        
 \hline 
        \citet{nguyen2018convolutional} & 146 & \textbullet & y & y & y & \textbullet & 2  \\
        
 \hline 
        \citet{miwa2016end} & 175 & \textbullet & y & y & y & \textbullet & 3  \\
        
 \hline 
        \citet{li2015multi} & 107 & y & y & y & \textbullet & y & 6  \\
        
 \hline 
        \citet{xu2015semantic} & 108 & \textbullet & y & y & \textbullet & y & 1  \\
        
 \hline 
        \citet{wang2016relation} & 102 & \textbullet & y & \textbullet & \textbullet & y & 1  \\
        
 \hline 
        \citet{hashimoto2013simple} & 64 & \textbullet & y & y & \textbullet & y & 1  \\
        
 \hline 
        \citet{zhang2015relation} & 68 & \textbullet & y & \textbullet & y & y & 2  \\
        
 \hline 
        \citet{vu2016combining} & 57 & \textbullet & y & y & \textbullet & y & 1  \\
        
 \hline 
        \citet{yin2017comparative} & 116 & \textbullet & n  & y & \textbullet & \textbullet & 7  \\
        
 \hline 
        \citet{yu2014factor} & 45 & y & y & y & y & y & 1  \\
        
 \hline 
        \citet{xu2016recurrent} & 54 & y  & y & y & \textbullet & \textbullet & 1  \\
        
 \hline 
        \citet{zhang2015bidirectional} & 51 & \textbullet & \textbullet & \textbullet & \textbullet & y & 1  \\
        
 \hline 
        \citet{nguyen2015combining} & 42 & \textbullet & y & y & \textbullet & y & 2  \\
        
 \hline 
        \citet{qin2016empirical} & 39 & \textbullet & \textbullet & y & y & y & 1  \\
        
 \hline 
        \citet{cai2016bidirectional} & 44 & \textbullet & y  & y & \textbullet & y & 1  \\
        
 \hline 
        \citet{sahu2016clinical} & 32 & \textbullet & y & y & y & y & 1  \\
        
 \hline 
        \citet{adel2016comparing} & 29 & y  & y & \textbullet & \textbullet & y & 1  \\
        
 \hline 
        \citet{zeng2015distant} & 190 & \textbullet & y & y & \textbullet & y & 1  \\
        
 \hline 
        \citet{xu2015classifying} & 171 & \textbullet & y & y & \textbullet & y & 1  \\
        
 \hline 
        \citet{zhang2018graphconvolution} & 3 & \textbullet & y & y & \textbullet & y & 2  \\
        
 \hline 
        \citet{omer2017zero} & 20 & y & y & y & \textbullet & y & 1  \\
        
 \hline 
        \citet{liu2016drug} & 48 & \textbullet & \textbullet & y & \textbullet & y & 1  \\
        
 \hline 
        \citet{zhao2016drug} & 41 & y  & y  & y & \textbullet & y & 1  \\
        
 \hline 
        \citet{ebrahimi2015chain} & 30 & \textbullet & \textbullet & \textbullet & \textbullet & \textbullet & 2  \\
        
 \hline 
        \citet{li2017neural} & 27 & y & y & y & y & y & 2  \\
        
 \hline 
        \citet{quan2016multichannel} & 23 & y & \textbullet & y & y & y & 2  \\
        
 \hline 
        \citet{sahu2018LSTM} & 13 & y  & y & y & \textbullet & y & 1  \\
        
 \hline 
        \citet{liu2016dependency} & 9 & \textbullet & \textbullet & y & \textbullet & y & 1  \\
        
 \hline 
        \citet{lim2018drugrecursive} & 4 & \textbullet & \textbullet & \textbullet & \textbullet & \textbullet & 1  \\
        
 \hline 
        \citet{zheng2017attention} & 12 & \textbullet & y & y & y & y & 1  \\
        
 \hline 
        \citet{wang2017dependency} & 5 & n  & y & y & \textbullet & y & 1  \\
        
 \hline 
        \citet{lim2018drug} & 1 & y  & y & y & y & y & 2  \\
        
 \hline 
        \citet{kavuluru2017extracting} & 8 & \textbullet & \textbullet & y & \textbullet & \textbullet & 1  \\
        
 \hline 
        \citet{huang2017SVMLSTM} & 4 & \textbullet & \textbullet & y & \textbullet & y & 1  \\
        
 \hline 
        \citet{juan2018extraction} & 1 & \textbullet & \textbullet & \textbullet & \textbullet & y & 1  \\
        
 \hline 
        \citet{lim2018chemicalgene} & 4 & y & \textbullet & y & \textbullet & y & 1  \\
        
 \hline 
        \citet{rotsztejn2018semeval} & 2 & \textbullet & \textbullet & y & y & y & 1  \\
        
 \hline 
        \citet{jin2018semeval} & 0 & \textbullet & y  & y & y & y & 1  \\
        
 \hline 
        \citet{sahu2016clinical} & 31 & \textbullet & y & y & y & y & 1  \\
        
 \hline 
        \citet{luo2017recurrent} & 21 & \textbullet & \textbullet & y & \textbullet & y & 1  \\
        
 \hline 
        \citet{lv2016clinical} & 15 & \textbullet & \textbullet & \textbullet & \textbullet & \textbullet & 1  \\
        
 \hline 
        \citet{jin2018semeval} & 14 & \textbullet & y & y & \textbullet & y & 1  \\
        
 \hline 
        \citet{chikka2018AHD} & 1 & y & \textbullet & y & \textbullet & \textbullet & 1  \\
        
 \hline 
        \citet{li2018classifying} & 0 & y  & \textbullet & y & y & y & 1  \\
        
 \hline 
        \citet{q2018multitask} & 0 & \textbullet & \textbullet & \textbullet & \textbullet & \textbullet & 5  \\
        
 \hline 
        \citet{suster2018revisiting} & 0 & y & \textbullet & y & \textbullet & y & 1  \\
        
 \hline 
        \citet{luo2017segment} & 16 & y & \textbullet & y & \textbullet & y & 1  \\
        
 \hline 
        \citet{he2018classifying} & 2 & \textbullet & \textbullet & y & \textbullet & y & 1  \\
        
 \hline 
        \citet{He2018ConvolutionalGR} & 0 & \textbullet & \textbullet & y & y & y & 2  \\
        
 \hline 
        \citet{nguyen2018convolutional} & 1 & \textbullet & y  & y & \textbullet & y & 1  \\
        
    \hline \hline
          
     \caption{Following are the columns in this table: \textbf{cite} = number of papers that cited the paper; \textbf{code} = whether code was publicly available (y for yes and \textbullet\ for no); \textbf{ablation} = whether an ablation study was performed; \textbf{hyperparam} = whether hyperparameter details were mentioned; \textbf{cross val} = whether cross validation details were mentioned; \textbf{word-embed} = whether information about word embeddings used was mentioned; \textbf{datasets} = number of datasets evaluated on}
     \label{tab:quant-review}
 \end{longtable}

\end{landscape}

\newpage
\clearpage
\twocolumn
\begin{landscape}
\section{Evaluation Metric Results on Test Data}
\label{appendix:evaluation-metric}

Each row represents a pre-processing, modeling technique or combination based on the additional experiments run on each dataset. Only test set results (as opposed to cross validation) are reported for ease of analysis. In all the tables, Baseline refers to the \texttt{CRCNN} model with original pre-processing and default hyperparameters for \texttt{semeval} and manual hyperparameters for the medical datasets (\texttt{ddi} and \texttt{i2b2}). 
The following short forms are used as row labels:

\textbf{B} = BERT-tokens 

\textbf{E} = ELMo 

\textbf{Ent Blind} = Entity Blinding 

\textbf{Piece Pool} = Piecewise Pooling 

  \begin{table}[h]
     \centering
     \begin{tabular}{l l l l l l l l}
     \hline \hline
     
     \diagbox[width=13em]{Technique}{Metric}      & \textbf{acc} & \textbf{micro-P} & \textbf{micro-R} & \textbf{micro-F1} & \textbf{macro-P}  & \textbf{macro-R} & \textbf{macro-F1}      \\ \hline

 \hline Baseline & 77.11 & 79.95 & 85.11 & 82.45 & 79.25 & 84.06 & 81.55  \\
 \hline Entity Blinding & 67.94 & 70.72 & 77.15 & 73.8 & 69.77 & 76.31 & 72.73  \\
 \hline Punct and Digit & 76.48 & 79.19 & 85.42 & 82.19 & 78.33 & 84.51 & 81.23  \\
 \hline Punct, Digit and Stop & 68.28 & 73.0 & 74.78 & 73.88 & 72.84 & 73.48 & 72.92  \\
 \hline NER Blinding & 77.25 & 79.3 & 86.03 & 82.53 & 78.49 & 85.13 & 81.63  \\
 \hline Piecewise pool & 77.0 & 79.54 & 85.55 & 82.44 & 78.86 & 84.71 & 81.59  \\
 \hline ELMo & 77.77 & 81.87 & 84.62 & 83.22 & 81.24 & 83.71 & 82.42  \\
 \hline BERT-CLS & 77.77 & 81.87 & 84.62 & 83.22 & 81.24 & 83.71 & 82.42  \\
 \hline BERT-tokens & 81.3 & 86.63 & 86.74 & 86.69 & 86.08 & 85.61 & 85.67  \\
    \hline \hline
    \end{tabular}
    \caption{Different Evaluation Metric results on test set of \texttt{semeval} dataset. Only test set results are reported for ease of analysis. Metric short forms used are \textbf{acc} = accuracy; \textbf{P} = precision; \textbf{R} = recall. }
    \label{tab:semeval-eval-metric}
 \end{table}
 
 \end{landscape}

\begin{landscape}
   \begin{table}[h]
     \centering
     \begin{tabular}{l | l l | l l | l l |  l l | l l | l l | l l}
     \hline \hline
     
     \multirow{2}{*}{\diagbox[width=10em]{Technique}{Metric}}      & \multicolumn{2}{c|}{\textbf{acc} } & \multicolumn{2}{c|}{\textbf{micro-P} } & \multicolumn{2}{c|}{\textbf{micro-R} } & \multicolumn{2}{c|}{\textbf{micro-F1} } & \multicolumn{2}{c|}{\textbf{macro-P} }  & \multicolumn{2}{c|}{\textbf{macro-R} } & \multicolumn{2}{c}{\textbf{macro-F1} }      \\ 
                                                &   Class & Detect & Class & Detect &   Class & Detect &   Class & Detect &   Class & Detect &   Class & Detect &   Class & Detect          \\ \hline

         \hline Baseline & 88.69 & 90.01 & 88.69 & 90.01 & 88.69 & 90.01 & 88.69 & 90.01 & 72.32 & 82.06 & 63.48 & 81.43 & 65.53 & 81.74  \\
         \hline Entity Blinding & 89.22 & 90.44 & 89.22 & 90.44 & 89.22 & 90.44 & 89.22 & 90.44 & 71.26 & 82.99 & 64.63 & 81.79 & 67.02 & 82.37  \\
         \hline Punct and Digit & 88.31 & 89.61 & 88.31 & 89.61 & 88.31 & 89.61 & 88.31 & 89.61 & 69.49 & 81.7 & 60.81 & 79.43 & 63.41 & 80.49  \\
         \hline Punct, Digit and Stop & 86.58 & 87.86 & 86.58 & 87.86 & 86.58 & 87.86 & 86.58 & 87.86 & 67.4 & 78.59 & 52.72 & 74.98 & 55.87 & 76.57  \\
         \hline NER Blinding & 86.18 & 88.74 & 86.18 & 88.74 & 86.18 & 88.74 & 86.18 & 88.74 & 59.13 & 79.9 & 55.93 & 78.24 & 57.22 & 79.03  \\
         \hline Piecewise pool & 88.14 & 89.54 & 88.14 & 89.54 & 88.14 & 89.54 & 88.14 & 89.54 & 70.49 & 81.39 & 60.38 & 79.91 & 63.01 & 80.62  \\
         \hline E & 89.76 & 90.97 & 89.76 & 90.97 & 89.76 & 90.97 & 89.76 & 90.97 & 73.41 & 84.36 & 63.65 & 81.9 & 66.63 & 83.05  \\
         \hline BERT-CLS & 87.84 & 89.05 & 87.84 & 89.05 & 87.84 & 89.05 & 87.84 & 89.05 & 68.2 & 80.51 & 59.31 & 78.84 & 61.3 & 79.63  \\
         \hline B & 91.31 & 92.72 & 91.31 & 92.72 & 91.31 & 92.72 & 91.31 & 92.72 & 77.66 & 87.34 & 69.27 & 85.78 & 71.97 & 86.53  \\
         \hline E + Entity Blinding & 89.97 & 91.18 & 89.97 & 91.18 & 89.97 & 91.18 & 89.97 & 91.18 & 72.44 & 84.42 & 66.41 & 83.06 & 68.69 & 83.72  \\
         \hline B + Entity Blinding & 90.93 & 92.15 & 90.93 & 92.15 & 90.93 & 92.15 & 90.93 & 92.15 & 76.79 & 86.57 & 63.39 & 84.26 & 70.66 & 85.35  \\
    \hline \hline
          
     \end{tabular}
    \caption{Different Evaluation Metric results on test set of \texttt{ddi} dataset. Only test set results are reported for ease of analysis. Metric short forms used are \textbf{acc} = accuracy; \textbf{P} = precision; \textbf{R} = recall. }
     \label{tab:ddi-eval-metric}
 \end{table}
\end{landscape}
 
 \begin{landscape}

\begin{table}[b]
     \centering
     \begin{tabular}{l | l l | l l | l l |  l l | l l | l l | l l}
     \hline \hline
     
     \multirow{2}{*}{\diagbox[width=13em]{Technique}{Metric}}      & \multicolumn{2}{c|}{\textbf{acc} } & \multicolumn{2}{c|}{\textbf{micro-P} } & \multicolumn{2}{c|}{\textbf{micro-R} } & \multicolumn{2}{c|}{\textbf{micro-F1} } & \multicolumn{2}{c|}{\textbf{macro-P} }  & \multicolumn{2}{c|}{\textbf{macro-R} } & \multicolumn{2}{c}{\textbf{macro-F1} }      \\ 
                                                &   Class & Detect & Class & Detect &   Class & Detect &   Class & Detect &   Class & Detect &   Class & Detect &   Class & Detect          \\ \hline

 \hline Baseline & 78.68 & 83.17 & 61.39 & 83.17 & 58.19 & 83.17 & 59.75 & 83.17 & 49.24 & 81.16 & 34.2 & 80.29 & 36.44 & 80.69  \\
 \hline Entity Blinding & 81.92 & 84.37 & 68.88 & 84.37 & 68.65 & 84.37 & 68.76 & 84.37 & 53.33 & 82.32 & 40.72 & 82.27 & 43.76 & 82.29  \\
 \hline Punct and Digit & 77.25 & 81.96 & 58.09 & 81.96 & 59.64 & 81.96 & 58.85 & 81.96 & 49.28 & 79.53 & 33.56 & 79.92 & 34.93 & 79.71  \\
 \hline Punct, Digit and Stop & 76.05 & 80.47 & 57.15 & 80.47 & 55.27 & 80.47 & 56.19 & 80.47 & 43.26 & 77.96 & 31.16 & 77.47 & 32.99 & 77.7  \\
 \hline NER Blinding & 75.12 & 81.61 & 52.58 & 81.61 & 48.42 & 81.61 & 50.41 & 81.61 & 39.44 & 79.45 & 26.3 & 78.17 & 29.15 & 78.73  \\
 \hline Piecewise pool & 78.63 & 83.69 & 59.41 & 83.69 & 62.37 & 83.69 & 60.85 & 83.69 & 46.16 & 81.41 & 35.77 & 82.17 & 36.44 & 81.76  \\
 \hline E & 80.4 & 84.54 & 64.56 & 84.54 & 61.86 & 84.54 & 63.18 & 84.54 & 59.28 & 82.69 & 36.17 & 81.97 & 38.1 & 82.31  \\
 \hline BERT-CLS & 76.94 & 81.91 & 57.66 & 81.91 & 55.95 & 81.91 & 56.79 & 81.91 & 49.88 & 76.61 & 32.4 & 79.15 & 34.05 & 79.37  \\
 \hline B & 80.79 & 84.91 & 64.92 & 84.91 & 61.4 & 84.91 & 63.11 & 84.91 & 58.05 & 83.08 & 36.8 & 82.1 & 39.31 & 82.55  \\
 \hline E + Entity Blinding & 83.62 & 86.17 & 72.43 & 86.17 & 68.6 & 86.17 & 70.46 & 86.17 & 60.79 & 84.65 & 40.11 & 83.67 & 42.99 & 84.13  \\
 \hline E + Piece Pool + Ent Blind & 83.46 & 86.14 & 71.11 & 86.14 & 70.14 & 86.14 & 70.62 & 86.14 & 54.87 & 84.37 & 42.41 & 84.13 & 44.43 & 84.25  \\
 \hline Ent Blind + Piece Pool & 82.72 & 85.44 & 69.49 & 85.44 & 69.98 & 85.44 & 69.73 & 85.44 & 48.82 & 83.49 & 41.97 & 83.61 & 42.89 & 83.55  \\
 \hline E + Piece Pool & 80.1 & 84.92 & 61.98 & 84.92 & 64.45 & 84.92 & 63.19 & 84.92 & 49.68 & 82.79 & 36.91 & 83.43 & 37.52 & 83.09  \\
 \hline B + Ent Blind & 83.27 & 85.66 & 71.52 & 85.66 & 69.63 & 85.66 & 70.56 & 85.66 & 55.62 & 83.9 & 38.82 & 83.44 & 41.83 & 83.66  \\
 \hline B + Ent Blind + Piece pool & 83.57 & 86.26 & 70.9 & 86.26 & 71.13 & 86.26 & 71.01 & 86.26 & 55.6 & 84.43 & 42.58 & 84.49 & 44.4 & 84.46  \\
 \hline B + Piece pool & 80.59 & 85.45 & 63.08 & 85.45 & 63.39 & 85.45 & 63.23 & 85.45 & 56.01 & 83.51 & 36.84 & 83.59 & 38.84 & 83.55  \\
    \hline \hline
    \end{tabular}
         \caption{Different Evaluation Metric results on test set of \texttt{i2b2} dataset. Only test set results are reported for ease of analysis. Metric short forms used are \textbf{acc} = accuracy; \textbf{P} = precision; \textbf{R} = recall. }
     \label{tab:i2b2-eval-metric}
 \end{table}

\end{landscape}

\end{document}